\documentclass{bmvc2k}

\newcommand{\op}[1]{\operatorname{#1}}
\newcommand{\bg}[1]{\boldsymbol{#1}} %Bold Greek letters
\newcommand{\bm}[1]{\mathbf{#1}} %Bold vectors and matrices
\newcommand{\vc}[3]{\overset{#2}{\underset{#3}{#1}}}

\newcommand\T{{\mathpalette\raiseT\intercal}}
\newcommand\raiseT[2]{%
\setbox0\hbox{$#1{#2}$}\raise\dp0\box0}

\usepackage{amsmath,amssymb}
\usepackage{booktabs}

\title{Higher-Order Implicit Fairing Networks for 3D Human Pose Estimation}

\addauthor{Jianning Quan}{jianning.quan@mail.concordia.ca}{1}
\addauthor{A. Ben Hamza}{hamza@ciise.concordia.ca}{1}
\addinstitution{
 Concordia University -- CIISE\\
 Montreal, QC, Canada
}

\runninghead{Quan, Hamza}{Higher-Order Implicit Fairing Networks}

\begin{document}

\maketitle

\begin{abstract}
Estimating a 3D human pose has proven to be a challenging task, primarily because of the complexity of the human body joints, occlusions, and variability in lighting conditions. In this paper, we introduce a higher-order graph convolutional framework with initial residual connections for 2D-to-3D pose estimation. Using multi-hop neighborhoods for node feature aggregation, our model is able to capture the long-range dependencies between body joints. Moreover, our approach leverages residual connections, which are integrated by design in our network architecture, ensuring that the learned feature representations retain important information from the initial features of the input layer as the network depth increases. Experiments and ablations studies conducted on two standard benchmarks demonstrate the effectiveness of our model, achieving superior performance over strong baseline methods for 3D human pose estimation.
\end{abstract}

%-------------------------------------------------------------------------

\section{Introduction}
The task of 3D human pose estimation is a fundamental problem in computer vision, robotics, and computer graphics. It refers to the process of predicting the positions of a person's joints (also known as keypoints or landmarks) in images or videos. Application domains of 3D human pose estimation are abundant, and range from activity recognition, surveillance and healthcare to games and sports.

Tremendous progress has been made in estimating 3D human pose from images or videos thanks to the rapid development of deep neural network solutions, which have been shown to achieve improved performance over classical approaches that use hand-crafted features. Most existing 3D pose estimation methods use an end-to-end pipeline~\cite{li20143d} or a two-stage pipeline~\cite{pavlakos2017coarse,sun2017compositional}. The former employs a deep neural network to regress 3D keypoints from images in an end-to-end fashion, whereas the latter is comprised of two main stages, which are usually decoupled from each other. Two-stage approaches for 3D pose estimation have shown great promise~\cite{zhou2017towards,martinez2017simple,yang20183d,fang2018learning,rayat2018exploiting,pavlakos2018ordinal,sharma2019monocular,ge20193d,pavllo20193d}, outperforming end-to-end models. This better performance is largely attributed to the fact that two-stage methods benefit from intermediate supervision provided, in part, by robust 2D pose detectors~\cite{pavllo20193d}. Martinez \textit{el al.}~\cite{martinez2017simple} design a simple fully connected network with residual connections for estimating 3D poses from 2D joint detections, outperforming systems trained end-to-end from raw pixels.

In recent years, there has been a surge of interest in the adoption of graph convolution networks (GCNs) for 3D pose estimation~\cite{zhao2019semantic,YujunCai:19,zou2020high}, achieving state-of-the-art performance. Much of this interest stems from the fact that a 2D human skeleton can naturally be represented as a graph whose nodes are body joints and edges are connections between neighboring joints. Zhao \textit{et al.}~\cite{zhao2019semantic} propose SemGCN, a semantic graph convolutional network, which learns to capture semantic information encoded in a given graph (i.e. local and global relations between nodes), yielding improved performance in 3D pose estimation while using a much smaller number of parameters. While GCN is powerful for learning on graph-structured data, it suffers, however, from the oversmoothing problem~\cite{Li:18}, where the learned node representations become indistinguishable due to repeated graph convolutions as the network depth increases. Several attempts have been made toward remedying this issue of oversmoothing~\cite{Keyulu:18,Klicpera:19,Lingxiao:20,chen2020simple}. Another issue with GCN is that its aggregation scheme uses one-hop neighbors, and hence lacks the ability to capture long-range dependencies. This issue can be mitigated by skipping connections during feature aggregation using, for example, the jumping knowledge networks~\cite{Keyulu:18} or by concatenating feature representations of multi-hop neighbors via sparsified neighborhood mixing (MixHop)~\cite{abu2019mixhop}, which leverages a graph convolutional layer that mixes powers of the adjacency matrix. Building on MixHop, Zou \textit{et al.}~\cite{zou2020high} propose a high-order GCN for 3D pose estimation, with the goal of capturing long-range dependencies between body joints.

To address the above issues, we introduce a higher-order graph convolutional framework for 3D pose estimation via implicit fairing on graphs~\cite{Desbrun:99}. We follow the two-stage paradigm by employing a state-of-art 2D pose detector, followed by a lifting network for predicting the 3D pose locations from the 2D predictions. The aggregation scheme of the proposed approach leverages residual connections to help alleviate the oversmoothing problem, and uses multi-hop neighborhoods to capture long-range dependencies between body joints. The main contributions of this work can be summarized as follows:

\begin{itemize}
\item We derive an implicit fairing network (IF-Net) with initial residual connection by iteratively solving the implicit fairing equation on graphs via Jacobi method.
\item We propose a higher-order implicit fairing network (HOIF-Net) for 3D human pose estimation by concatenating feature representations from multi-hop neighborhoods, with the aim to capture long-range dependencies.
\item We demonstrate through experiments and ablation studies that our proposed model achieves state-of-the-art performance in comparison with strong baselines.
\end{itemize}

\section{Related work}
\noindent\textbf{Graph Convolution Networks.}\quad GCNs have recently become the de facto model for learning representations on graphs. However, GCNs are prone to oversmooting as the network depth increases, and also fail to capture important dependencies between distant nodes. To circumvent these limitations, a plethora of GCN variants have been proposed, including jumping knowledge networks (JK-Nets)~\cite{Keyulu:18}, graph convolutional networks with initial residual connection and identity mapping (GCNII)~\cite{chen2020simple}, and higher-order graph convolutional architectures via MixHop~\cite{abu2019mixhop}. The latter learns neighborhood mixing relationships by repeatedly mixing feature representations of neighbors at various distances through powers of the graph adjacency matrix, while requiring no additional memory or computational complexity.

\medskip\noindent\textbf{3D Human Pose Estimation.}\quad Most approaches to 3D human pose estimation can generally be classified into two main categories, namely single-stage and two-stage models, with the former using an end-to-end pipeline to predict 3D poses from images; and the latter using a two-stage pipeline, in which 2D joint locations are first extracted using a 2D pose detector and then a lifting network is employed to regress 3D poses from 2D detections. Our approach falls under the category of two-stage models~\cite{zhou2017towards,martinez2017simple,yang20183d,fang2018learning,rayat2018exploiting,pavlakos2018ordinal,sharma2019monocular,ge20193d,pavllo20193d,zhao2019semantic,YujunCai:19,HaiCi:2019,Kenkun:2020,zou2020high}. Zou \textit{et al.}~\cite{zou2020high} design a high-order GCN model for 3D pose estimation based on MixHop in a bid to capture long-range dependencies between distant body joints using a network architecture comprised of a residual block repeated several times similar to the network design of Martinez \textit{et al.}~\cite{martinez2017simple}. However, the model inherits the oversmoothing issue of GCNs, where repeated graph convolutions make learned node embeddings indistinguishable; thereby, resulting in performance drop. By contrast, our proposed network architecture has residual connections integrated by design, and hence is able to alleviate the oversmoothing problem. This is in line with existing approaches such as jumping knowledge networks~\cite{Keyulu:18} and graph convolutional networks with initial residual and identity mapping~\cite{chen2020simple}. In addition, we use a scaled, learnable weight matrix with a layer-dependent scale factor in an effort to ensure that the weight decay adaptively increases as more layers are added~\cite{chen2020simple}.

\section{Preliminaries and Problem Statement}
\noindent\textbf{Basic Notions.}\quad Consider a graph $\mathcal{G}=(\mathcal{V},\mathcal{E})$, where $\mathcal{V}=\{1,\ldots,N\}$ is the set of $N$ nodes (e.g., body joints) and $\mathcal{E}\subseteq \mathcal{V}\times\mathcal{V}$ is the set of edges (e.g., connections between two body joints). Let $\bm{A}$ be an $N\times N$ adjacency matrix whose $(i,j)$-th entry is equal to the weight of the edge between neighboring nodes $i$ and $j$, and 0 otherwise. We denote by $\tilde{\bm{A}}=\bm{A}+\bm{I}$ the adjacency matrix with self-added loops, where $\bm{I}$ is the identity matrix. We also denote by $\bm{X}=(\bm{x}_{1},...,\bm{x}_{N})^{\T}$ an $N\times F$ feature matrix of node attributes, where $\bm{x}_{i}$ is an $F$-dimensional row vector for node $i$. We define the normalized Laplacian matrix as follows:
\begin{equation}
\bm{L}=\bm{I}-\tilde{\bm{D}}^{-\frac{1}{2}}\tilde{\bm{A}}\tilde{\bm{D}}^{-\frac{1}{2}},
\end{equation}
where $\tilde{\bm{D}}=\op{diag}(\tilde{\bm{A}}\bm{1})$ is the diagonal degree matrix, and $\bm{1}$ is an $N$-dimensional vector of all ones. The Laplacian matrix admits an eigendecomposition given by $\bm{L}=\bm{U}\bg{\Lambda}\bm{U}^{\T}$, where $\bm{U}$ is an orthonormal matrix whose columns constitute an orthonormal basis of eigenvectors and $\bg{\Lambda}$ is a diagonal matrix comprised of the corresponding eigenvalues.

\medskip\noindent\textbf{Graph Convolutional Networks (GCNs).}\quad Given an input feature matrix $\bm{H}^{(\ell)}\in\mathbb{R}^{N\times F_{\ell}}$ of the $\ell$-th layer with $F_{\ell}$ feature maps, the output feature matrix $\bm{H}^{(\ell+1)}$ of GCN is obtained by applying the following layer-wise propagation rule:
\begin{equation}
\bm{H}^{(\ell+1)}=\sigma(\tilde{\bm{D}}^{-\frac{1}{2}}\tilde{\bm{A}}\tilde{\bm{D}}^{-\frac{1}{2}}\bm{H}^{(\ell)}\bm{W}^{(\ell)}),\quad \ell=0,\dots,L-1,
\label{Eq:AGCNprop}
\end{equation}
which is basically a node embedding transformation that projects $\bm{H}^{(\ell)}$ into a trainable weight matrix $\bm{W}^{(\ell)}\in\mathbb{R}^{F_{\ell}\times F_{\ell+1}}$ with $F_{\ell +1}$ feature maps, followed by an activation function $\sigma(\cdot)$ such as $\text{ReLU}(\cdot)=\max(0,\cdot)$. The input of the first layer is the initial feature matrix $\bm{H}^{(0)}=\bm{X}$.

\medskip\noindent\textbf{Jacobi Method.}\quad The Jacobi method~\cite{Saad:03} is an iterative approach for solving a matrix equation $\bm{M}\bm{x}=\bm{b}$, where the square matrix $\bm{M}$ has no zeros along its main diagonal, by first decomposing $\bm{M}$ into a diagonal component and an off-diagonal component, i.e.
\begin{equation}
\bm{M}=\op{diag}(\bm{M})+\op{off}(\bm{M}).
\end{equation}
Then, the solution of the matrix equation $\bm{M}\bm{x}=\bm{b}$ is obtained iteratively as follows:
\begin{equation}
\bm{x}^{(t+1)} = \op{diag}(\bm{M})^{-1}(\bm{b}-\op{off}(\bm{M})\bm{x}^{(t)}),
\end{equation}
where $\bm{x}^{(t)}$ and $\bm{x}^{(t+1)}$ are the $t$-th and $(t+1)$-th iterations of $\bm{x}$, respectively.

\medskip\noindent\textbf{Problem Statement.}\quad Let $\mathcal{D}_{l}=\{(\bm{x}_{i},\bm{y}_{i})\}_{i=1}^{N}$ be a training set of 2D joint positions $\bm{X}=(\bm{x}_{1},\dots,\bm{x}_{N})^{\T}\in\mathbb{R}^{N\times 2}$ and their associated 3D joint positions $\bm{Y}=(\bm{y}_{1},\dots,\bm{y}_{N})^{\T}\in\mathbb{R}^{N\times 3}$. The goal of 3D human pose estimation is to learn the parameters $\bm{w}$ of a regression model $f: \bm{X}\to\bm{Y}$ by minimizing the following loss function
\begin{equation}
\bm{w}^{*}=\arg\min_{\bm{w}}\frac{1}{N}\sum_{i=1}^{N}\mathcal{L}(f(\bm{x}_{i}),\bm{y}_{i}).
\end{equation}
Since the 3D human pose estimation task is a regression problem, we train the model to minimize the mean squared error as a loss function.

\section{Proposed Method}
\subsection{Implicit Fairing on Graphs}
Applying a spectral graph filter with transfer function $h$ on the graph signal $\bm{X}$ yields a filtered graph signal $\bm{H}$ given by
\begin{equation}
\bm{H}=h(\bm{L})\bm{X}=\bm{U}h(\bg{\Lambda})\bm{U}^{\T}\bm{X}.
\label{Eq:SGF}
\end{equation}
Spectral graph filters are usually approximated using Chebyshev polynomials~\cite{Taubin:96,Hammond:11,Defferrard:16} or rational polynomials~\cite{Levie:18,Wijesinghe:19}. The implicit fairing method, which uses implicit integration of a diffusion process for graph filtering, has shown to allow for both efficiency and stability~\cite{Desbrun:99}. The implicit fairing filter is an infinite impulse response filter whose transfer function is given by $h_{s}(\lambda)=1/(1+s\lambda)$, where $s$ is a positive parameter. Substituting $h$ with $h_s$ in Eq.~\eqref{Eq:SGF}, we obtain
\begin{equation}
\bm{H}=(\bm{I}+s\bm{L})^{-1}\bm{X},
\end{equation}
where $\bm{I}+s\bm{L}$ is a symmetric positive definite matrix (all its eigenvalues are positive), and hence admits an inverse. Therefore, performing graph filtering with implicit fairing is equivalent to solving the following sparse linear system:
\begin{equation}
(\bm{I}+s\bm{L})\bm{H}=\bm{X}.
\label{Eq:IF}
\end{equation}

\subsection{Iterative Solution}
The implicit fairing equation \eqref{Eq:IF} can be solved iteratively using Jacobi's method, which uses matrix splitting. We can split the matrix $\bm{I}+s\bm{L}$ into the sum of a diagonal matrix and an off-diagonal matrix as follows:
\begin{equation}
\bm{I}+s\bm{L}=\op{diag}(\bm{I}+s\bm{L})+\op{off}(\bm{I}+s\bm{L}),
\end{equation}
where
$$\op{diag}(\bm{I}+s\bm{L})=(1+s)\bm{I} \quad \text{and}\quad \op{off}(\bm{I}+s\bm{L})=-s\tilde{\bm{D}}^{-\frac{1}{2}}\tilde{\bm{A}}\tilde{\bm{D}}^{-\frac{1}{2}}.$$
Hence, the iterative solution of the implicit fairing equation is given by
\begin{equation}
\begin{split}
\bm{H}^{(t+1)}
&=-(\op{diag}(\bm{I}+s\bm{L}))^{-1}\op{off}(\bm{I}+s\bm{L})\bm{H}^{(t)} + (\op{diag}(\bm{I}+s\bm{L}))^{-1}\bm{X}\\
&=(s/(1+s))\tilde{\bm{D}}^{-\frac{1}{2}}\tilde{\bm{A}}\tilde{\bm{D}}^{-\frac{1}{2}}\bm{H}^{(t)}+(1/(1+s))\bm{X},
\end{split}
\label{Eq:ISJ}
\end{equation}
which can be rewritten as
\begin{equation}
\bm{H}^{(t+1)}=(1-\alpha)\tilde{\bm{D}}^{-\frac{1}{2}}\tilde{\bm{A}}\tilde{\bm{D}}^{-\frac{1}{2}}\bm{H}^{(t)}+\alpha\bm{X},
\label{Eq:Jacobi}
\end{equation}
where the hyperparameter $\alpha=1/(1+s)\in (0,1)$, and $\bm{H}^{(t)}$ is the $t$-th iteration of $\bm{H}$.

\subsection{Implicit Fairing Network}
Inspired by the Jacobi iterative solution \eqref{Eq:Jacobi} of the implicit fairing equation, we propose a multi-layer implicit fairing network (IF-Net) with the following layer-wise propagation rule:
\begin{equation}
\bm{H}^{(\ell+1)}=\sigma(((1-\alpha)\bm{S}\bm{H}^{(\ell)}+\alpha\bm{X})\tilde{\bm{W}}^{(\ell)}),
\label{Eq:IFNet}
\end{equation}
where $\bm{S}=\tilde{\bm{D}}^{-\frac{1}{2}}\tilde{\bm{A}}\tilde{\bm{D}}^{-\frac{1}{2}}$ is the normalized adjacency matrix with self-added loops, and $\tilde{\bm{W}}^{(\ell)}=\beta_{\ell}\bm{W}^{(\ell)}$ is a scaled, learnable weight matrix with a layer-dependent scale factor defined as $\beta_{\ell}=\log(1+\beta/(1+\ell))$, which ensures that the decay of the weight matrix increases in tandem with the network depth~\cite{chen2020simple}.
%The hyperparameters $\alpha$ and $\beta$ are selected via grid search with cross-validation.

\subsection{Higher-Order Implicit Fairing Network}
\begin{figure}[!htb]
\setlength{\tabcolsep}{1em}
\centering
\begin{tabular}{cc}
\includegraphics[scale=.36]{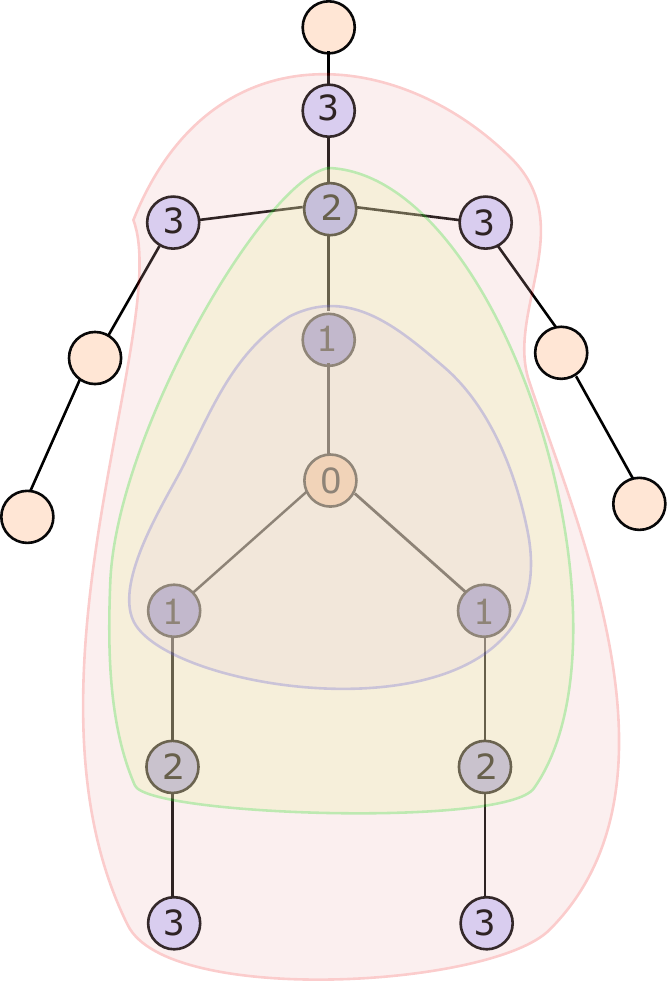} &
\includegraphics[scale=.48]{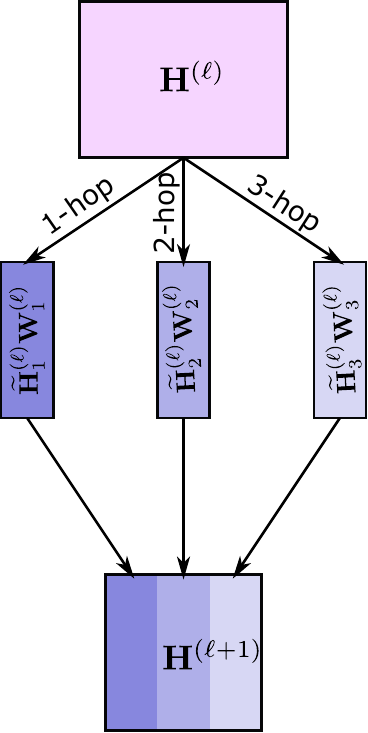}
\end{tabular}
\caption{Illustration of HOIF-Net feature concatenation for $K=3$.}
\label{Fig:humangraph2}
\end{figure}

Using the feature diffusion rule of GCN is tantamount to applying a weighted sum of the features of neighboring nodes normalized by their degrees, which essentially performs Laplacian smoothing on the graph~\cite{Li:18}, and hence leads to oversmoothing. Also, the aggregation scheme of GCN uses 1-hop neighbors, and hence lacks the ability to capture long-range dependencies. To circumvent these issues, we define a higher-order implicit fairing network (HOIF-Net) with the following layer-wise propagation rule:
\begin{equation}
\bm{H}^{(\ell+1)}=\sigma(\vc{\parallel}{K}{k=1}\tilde{\bm{H}}_{k}^{(\ell)}\tilde{\bm{W}}_{k}^{(\ell)}),
\label{Eq:IFNet}
\end{equation}
where
\begin{equation}
\tilde{\bm{H}}_{k}^{(\ell)}= (1-\alpha)\bm{S}^{k}\bm{H}^{(\ell)}+\alpha\bm{X},
\label{Eq:transf}
\end{equation}
and $\bm{S}^{k}$ is the $k$-th power of the normalized adjacency matrix with self-added loops. Each $(i,j)$-th entry of $\bm{S}^{k}$ counts the number of walks of length $k$ between nodes $i$ and $j$. For example, the $(i,j)$-th entry of $\bm{S}^{2}$ gives the number of common neighbors of nodes $i$ and $j$. The learnable weight matrix $\tilde{\bm{W}}_{k}^{(\ell)}$ is associated to the the node feature representation $\tilde{\bm{H}}_{k}^{(\ell)}$, and $\parallel$ denotes concatenation. For each $k$-hop neighborhood, the node feature representation $\tilde{\bm{H}}_{k}^{(\ell)}$ given by Eq.~\eqref{Eq:transf} is a weighted sum of the transformed feature matrix $\bm{S}^{k}\bm{H}^{(\ell)}$ for the $\ell$-layer and the initial feature matrix $\bm{X}$. Intuitively, the transformation $\bm{S}^{k}\bm{H}^{(\ell)}$ yields a smooth hidden representation, and hence encourages similar predictions among $k$-hop neighboring nodes. The weighting factor $\alpha$ represents the weight assigned to the initial feature information that needs to be carried over, as the number of layers increase. Figure~\ref{Fig:humangraph2} shows an illustration of the layer-wise propagation rule of HOIF-Net when $K=3$. Long-range dependencies between body joints are captured by high-order graph convolutions, which take into account distant neighbors when updating the learned node features. Note that HOIF-Net uses residual connections between the initial feature matrix and each hidden layer. Residual connections not only allow the model to carry over information from the initial node attributes, but also help facilitate training of multi-layer networks.

\begin{figure*}[!htb]
\centering
\includegraphics[scale=.6]{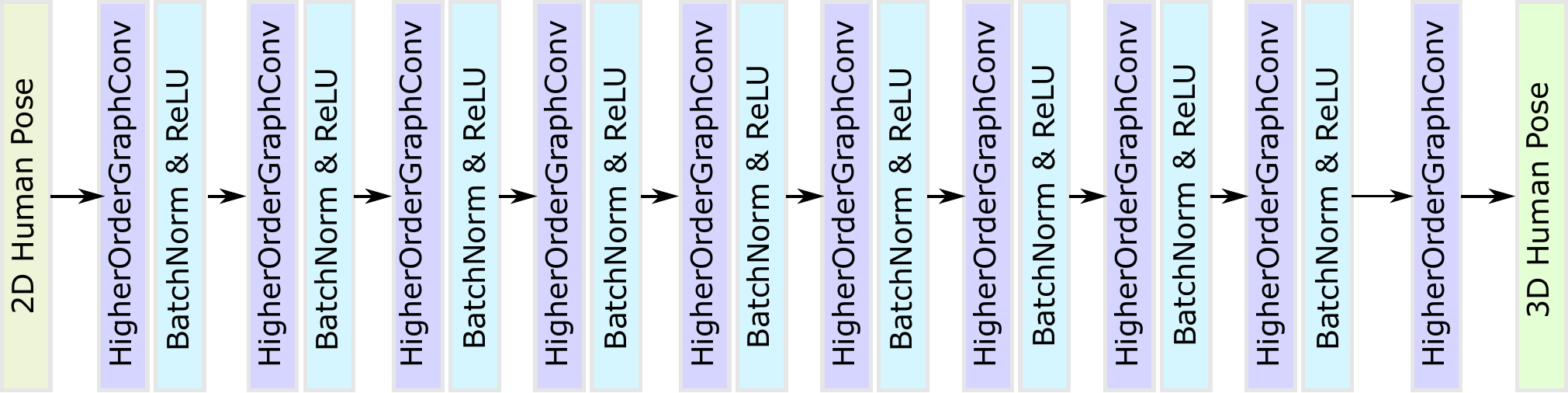}
\caption{Overview of the proposed network architecture for 3D pose estimation. Our model takes 2D pose coordinates (17 joints) as input and generates 3D pose predictions (17 joints) as output. We use ten higher-order graph convolutional layers, each of which is followed by batch normalization and ReLU activation function, except the last convolutional layer.}
\label{Fig:model}
\end{figure*}

\medskip\noindent\textbf{Model Architecture.}\quad The architecture of our proposed model for 3D human pose estimation is illustrated in Figure~\ref{Fig:model}. The input consists of 2D keypoints generated via a 2D pose detector. The generated output of the proposed model consists of predicted 3D pose coordinates. We use higher-order graph convolutional layers defined by the layer-wise propagation rule of HOIF-Net to capture long-range structural information between body joints.

\medskip\noindent\textbf{Model Prediction.}\quad The output of the last higher-order graph convolutional layer of HOIF-Net contains the final output node embeddings, which are given by
\begin{equation}
\hat{\bm{Y}}=(\hat{\bm{y}}_{1},\dots,\hat{\bm{y}}_{N})^{\T}\in\mathbb{R}^{N\times 3},
\end{equation}
where $\hat{\bm{y}}_{i}$ is a three-dimensional raw vector of predicted 3D pose coordinates.

\medskip\noindent\textbf{Model Training.}\quad The parameters (i.e. weight matrices for different layers) of the proposed HOIF-Net model for 3D human pose estimation are learned by minimizing the loss function
\begin{equation}
\mathcal{L} =\frac{1}{N}\sum_{i=1}^{N}\Vert\bm{y}_{i}-\hat{\bm{y}}_{i}\Vert_{2}^{2},
\end{equation}
which is the mean squared error between the 3D ground truth poses $\bm{y}_{i}$ and estimated 3D joint poses $\hat{\bm{y}}_{i}$ over a training set consisting of $N$ human poses.

\section{Experiments}
\subsection{Experimental Setup}
\medskip\noindent\textbf{Datasets.}\quad We perform quantitative and qualitative evaluations on two standard, large-scale benchmark datasets: Human 3.6M and MPI-INF-3DHP. The Human 3.6M dataset~\cite{ionescu2013human3} contains 3.6 million 3D human poses for 11 professional actors and corresponding images captured by a high-speed motion capture system with four different cameras. Each actor performs 15 actions (scenarios), including directions, discussion, eating, greeting, talking on the phone and so on. For data preprocessing, we apply standard normalization to the 2D and 3D poses before feeding the data to the model in line with previous work~\cite{zou2020high, martinez2017simple}. For the MPI-INF-3DHP dataset~\cite{Dushyant:2017}, there are 8 actors performing 8 activities each. These activities range from walking and sitting to complex exercise poses and dynamic actions.

\medskip\noindent\textbf{Evaluation Protocols and Metrics.}\quad For the Human 3.6M benchmark, there are two commonly used evaluation protocols, referred to as Protocol \#1 and Protocol \#2. Both protocols use 5 subjects (S1, S5, S6, S7, S8) for training and 2 subjects (S9, S11) for testing. Under Protocol \#1, we report the mean per joint position error (MPJPE), which computes the average Euclidean distance between the predicted 3D joint positions and ground truth after the alignment of the root joint (central
hip). Under Protocol \#2, we report the Procrustes-aligned mean per joint position error (PA-MPJPE), where MPJPE is computed after rigid alignment of the prediction with respect to the ground truth. Both error metrics are measured in millimeters, and lower values indicate better performance. For MPI-INF-3DHP, we adopt two commonly-used evaluation metrics, namely Percentage of Correct Keypoints (PCK) under 150mm and the Area Under the Curve (AUC), following previous works~\cite{yang20183d,pavlakos2018ordinal,Habibie:19}. Higher values of PCK and AUC indicate better performance.

\medskip\noindent\textbf{Implementation Details.}\quad We train our model for 50 epochs using the Adam optimizer with a learning rate of 0.001. We set the decay factor to 0.96 per 100,000 steps, and the batch size to 64. We also set the hyperparameters $\alpha$ and $\beta$ to 0.2 and 0.5, respectively, via grid search with cross-validation on the training set. To extract 2D keypoints from input images and following common practices in previous work~\cite{pavllo20193d,zou2020high}, we employ the cascaded pyramid network (CPN)~\cite{chen2018cascaded}, which uses bounding boxes obtained by Mask R-CNN~\cite{KHe:17}. For $K$-hop feature concatenation, we set the value of $K$ to 3, as illustrated in Figure~\ref{Fig:humangraph2}.

\subsection{Results and Analysis}
\noindent\textbf{Quantitative Results.}\quad In Tables~\ref{Tab:Result1} and~\ref{Tab:Result2}, we summarize the performance comparison results of our HOIF-Net model and various state-of-the-art methods for 3D pose estimation. As can be seen, our model performs the best in most of the actions and also on average under both Protocol \#1 and Protocol \#2, indicating that HOIF-Net is very competitive. Under Protocol \#1, Table~\ref{Tab:Result1} shows that HOIF-Net performs better than high-order GCN~\cite{zou2020high} on 14 out of 15 actions, yielding an error reduction of approximately 1.44\% on average over high-order GCN. Moreover, our model outperforms semGCN~\cite{zhao2019semantic} by a relative improvement of 4.86\% on average. Under Protocol \#2, Table~\ref{Tab:Result2} shows that our model performs better than high-order GCN with 1.83\% error reduction on average, and also achieves better performance on 11 out of 15 actions.

\begin{table*}[!htb]
\caption{Performance comparison of our model and baseline methods using MPJPE (in millimeters) between the ground truth and estimated pose on Human3.6M under Protocol \#1. The last column report the average errors, and boldface numbers indicate the best 3D pose estimation performance.}
\footnotesize
\setlength{\tabcolsep}{.4pt}
\medskip
\centering
\begin{tabular}{l*{17}{c}}
\toprule
& \multicolumn{15}{c}{Action}\\
\cmidrule(lr){2-16}
Method & Dire. & Disc. &  Eat & Greet & Phone & Photo &  Pose & Purch. & Sit & SitD. & Smoke & Wait & WalkD. & Walk & WalkT. & Avg.\\
\midrule
Martinez \textit{et al.}~\cite{martinez2017simple} &  51.8 &56.2 &58.1 &59.0 &69.5& 78.4 &55.2 &58.1 &74.0 &94.6 &62.3 &59.1& 65.1& 49.5& 52.4 &62.9\\
Sun \textit{et al.}~\cite{sun2017compositional} &52.8& 54.8& 54.2& 54.3 &61.8 &67.2& 53.1& 53.6 &71.7 &86.7 &61.5 &53.4& 61.6 &47.1& 53.4 &59.1\\
Yang \textit{et al.}~\cite{yang20183d} & 51.5& 58.9& \textbf{50.4} &57.0& 62.1& 65.4 &49.8& 52.7& 69.2& 85.2& 57.4& 58.4& 43.6& 60.1& 47.7& 58.6\\
Fang \textit{et al.}~\cite{fang2018learning}& 50.1 &54.3& 57.0& 57.1& 66.6& 73.3& 53.4& 55.7& 72.8& 88.6& 60.3 &57.7& 62.7& 47.5 &50.6& 60.4\\
Hossain \& Little~\cite{rayat2018exploiting}  & 48.4 & \textbf{50.7} & 57.2 & 55.2 & 63.1 & 72.6 & 53.0 & 51.7 & 66.1 & 80.9 & 59.0 & 57.3 & 62.4 & 46.6 & 49.6 & 58.3\\
Pavlakos \textit{et al.}~\cite{pavlakos2018ordinal} & 48.5& 54.4& 54.4& \textbf{52.0} &59.4 &65.3 &49.9& 52.9& 65.8 &71.1& 56.6& 52.9& 60.9& 44.7& 47.8& 56.2\\
Sharma \textit{et al.}~\cite{sharma2019monocular} & 48.6 &54.5& 54.2& 55.7& 62.2& 72.0& 50.5& 54.3& 70.0& 78.3 &58.1& 55.4& 61.4& 45.2& 49.7& 58.0\\
Zhao \textit{et al.}~\cite{zhao2019semantic} & 47.3& 60.7& 51.4 &60.5& 61.1& \textbf{49.9}& \textbf{47.3}& 68.1 &86.2& \textbf{55.0}& 67.8& 61.0& \textbf{42.1}& 60.6& 45.3& 57.6\\
Li \textit{et al.}~\cite{ChenLiLee:2020} (BH) & 62.0 & 69.7 & 64.3 & 73.6 & 75.1 & 84.8 & 68.7 & 75.0 & 81.2 & 104.3 & 70.2 & 72.0 & 75.0 & 67.0 & 69.0 & 73.9\\
Banik \textit{et al.}~\cite{Banik:2021} & 51.0 & 55.3 & 54.0 & 54.6 & 62.4 & 76.0 & 51.6 & 52.7 & 79.3 & 87.1 & 58.4 & 56.0 & 61.8 & 48.1 & \textbf{44.1} & 59.5\\
Xu \textit{et al.}~\cite{YuanluXu:2021} & 47.1 & 52.8 & 54.2 & 54.9 & 63.8 & 72.5 & 51.7 & 54.3 & 70.9 & 85.0 & 58.7 & 54.9 & 59.7 & 43.8 & 47.1 & 58.1\\
Zou \textit{et al.}~\cite{zou2020high} & 49.0& 54.5& 52.3& 53.6& 59.2 &71.6& 49.6& 49.8 &66.0 &75.5 &55.1 &53.8& 58.5& \textbf{40.9}& 45.4 &55.6\\
\midrule
Ours &\textbf{47.0} & 53.7 & 50.9 & 52.4&\textbf{57.8} &71.3&50.2 &\textbf{49.1} &\textbf{63.5} &76.3  &\textbf{54.1}&\textbf{51.6} &56.5 &41.7 &45.3 &\textbf{54.8} \\
\bottomrule
\end{tabular}
\label{Tab:Result1}
\end{table*}

\begin{table*}[!htb]
\caption{Performance comparison of our model and baseline methods using PA-MPJPE between the ground truth and estimated pose on Human3.6M under Protocol \#2.}
\footnotesize
\setlength{\tabcolsep}{.4pt}
\medskip
\centering
\begin{tabular}{l*{17}{c}}
\toprule
& \multicolumn{15}{c}{Action}\\
\cmidrule(lr){2-16}
Method & Dire. & Disc. &  Eat & Greet & Phone & Photo &  Pose & Purch. & Sit & SitD. & Smoke & Wait & WalkD. & Walk & WalkT. & Avg.\\
\midrule
Pavlakos \textit{et al.}~\cite{pavlakos2017coarse} & 47.5 &50.5 &48.3& 49.3& 50.7 &55.2 &46.1 &48.0& 61.1& 78.1 &51.1& 48.3& 52.9& 41.5& 46.4 &51.9 \\
Zhou \textit{et al.}~\cite{zhou2017towards} & 47.9& 48.8 &52.7& 55.0& 56.8& 49.0 &45.5 &60.8& 81.1 &\textbf{53.7}& 65.5& 51.6& 50.4 &54.8 &55.9& 55.3\\
Martinez \textit{et al.}~\cite{martinez2017simple} & 39.5 &43.2 &46.4 &47.0 &51.0& 56.0 &41.4& 40.6 &56.5& 69.4& 49.2& 45.0& 49.5& 38.0 &43.1 &47.7\\
Sun \textit{et al.}~\cite{sun2017compositional} & 42.1& 44.3& 45.0 &45.4 &51.5 &53.0 &43.2& 41.3& 59.3 &73.3& 51.0& 44.0& 48.0& 38.3& 44.8& 48.3\\
Fang \textit{et al.}~\cite{fang2018learning} & 38.2& 41.7& 43.7& 44.9& 48.5 &55.3& 40.2& 38.2& 54.5 &64.4& 47.2 &44.3& 47.3& 36.7& 41.7& 45.7\\
Hossain \& Little~\cite{rayat2018exploiting}  & \textbf{35.7} &\textbf{39.3}& 44.6 &43.0& 47.2& 54.0& 38.3 &37.5 &51.6 &61.3& 46.5& 41.4 &47.3 &34.2 &39.4& 44.1\\
Lee \textit{et al.}~\cite{Lee2018LSTM}  & 38.0 & 39.3 & 46.3 & 44.4 & 49.0 & 55.1 & 40.2 & 41.1 & 53.2 & 68.9 & 51.0 & \textbf{39.1} & \textbf{33.9} & 56.4 & 38.5 & 46.2 \\
Li \textit{et al.}~\cite{ChenLiLee:2020} (BH) & 38.5 & 41.7 & 39.6 & 45.2 & 45.8 & \textbf{46.5} & 37.8 & 42.7 & 52.4 & 62.9 & 45.3 & 40.9 & 45.3 & 38.6 & 38.4 & 44.3\\
Banik \textit{et al.}~\cite{Banik:2021} & 38.4 & 43.1 & 42.9 & 44.0 & 47.8 & 56.0 & 39.3 & 39.8 & 61.8 & 67.1 & 46.1 & 43.4 & 48.4 & 40.7 & \textbf{35.1} & 46.4\\
Xu \textit{et al.}~\cite{YuanluXu:2021} & 36.7 & 39.5 & 41.5 & 42.6 & 46.9 & 53.5 & 38.2 & \textbf{36.5} & 52.1 & 61.5 & 45.0 & 42.7 & 45.2 & 35.3 & 40.2 & 43.8\\
Zou \textit{et al.}~\cite{zou2020high} &38.6 &42.8& 41.8 &43.4 &44.6& 52.9& \textbf{37.5}& 38.6 &53.3 &60.0& 44.4& 40.9& 46.9 &32.2 &37.9 &43.7\\
\midrule
Ours &36.9 &42.1&\textbf{40.3} &\textbf{42.1} &\textbf{43.7} &52.7&37.9 &37.7 &\textbf{51.5} &60.3  &\textbf{43.9}&\textbf{39.4} & 45.4 & \textbf{31.9} & 37.8 & \textbf{42.9} \\
\bottomrule
\end{tabular}
\label{Tab:Result2}
\end{table*}

Table~\ref{Tab:MPI} reports the quantitative comparison results of HOIF-Net and baseline methods on the MPI-INF-3DHP dataset. As can be seen, our method achieves the best performance on all evaluation metrics.

\begin{table}[!htb]
\caption{Performance comparison of our model and baseline methods on the MPI-INF-3DHP dataset using PCK and AUC as evaluation metrics. Higher values in boldface indicate the best performance.}
\small
\setlength{\tabcolsep}{2.5pt}
\medskip
\centering
\begin{tabular}{lcc}
\toprule
Method & PCK & AUC\\
\midrule
Yang \textit{et al.}~\cite{yang20183d} & 69.0 & 32.0 \\
Pavlakos \textit{et al.}~\cite{pavlakos2018ordinal}  & 71.9 & 35.3 \\
Habibie \textit{et al.}~\cite{Habibie:19}  & 70.4 & 36.0 \\
\midrule
Ours & \textbf{72.8} &\textbf{36.5} \\
\bottomrule
\end{tabular}
\label{Tab:MPI}
\end{table}

\begin{figure*}[!htb]
\centering
\setlength{\tabcolsep}{7pt}
\begin{tabular}{cc}
\hspace*{.25in} Input\hspace*{.2in} Prediction\hspace*{.1in} Ground Truth & \hspace*{.25in} Input\hspace*{.25in} Prediction\hspace*{.1in} Ground Truth \\
\includegraphics[width=2.2in]{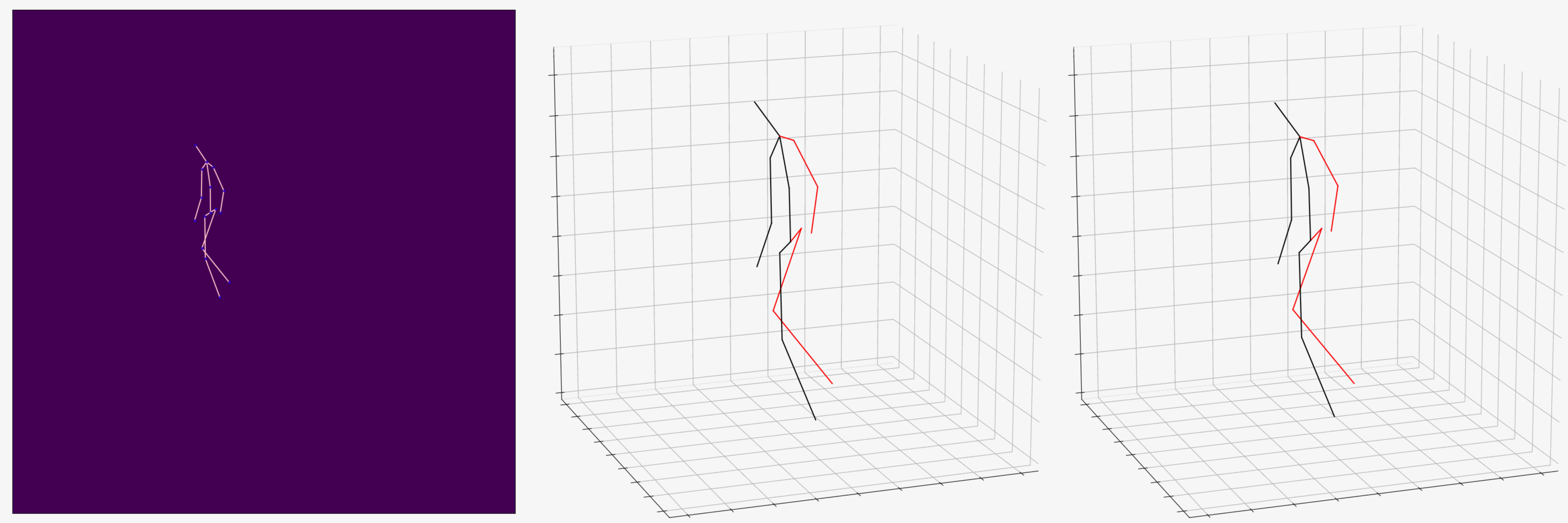} & \includegraphics[width=2.2in]{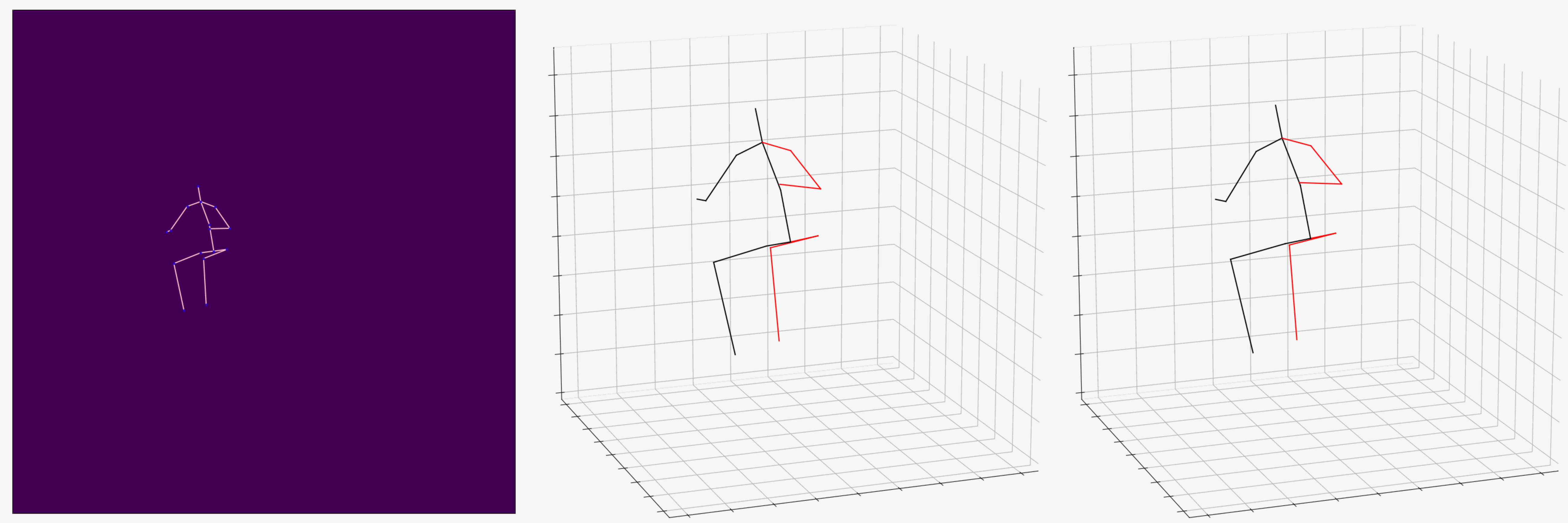}\\
\includegraphics[width=2.2in]{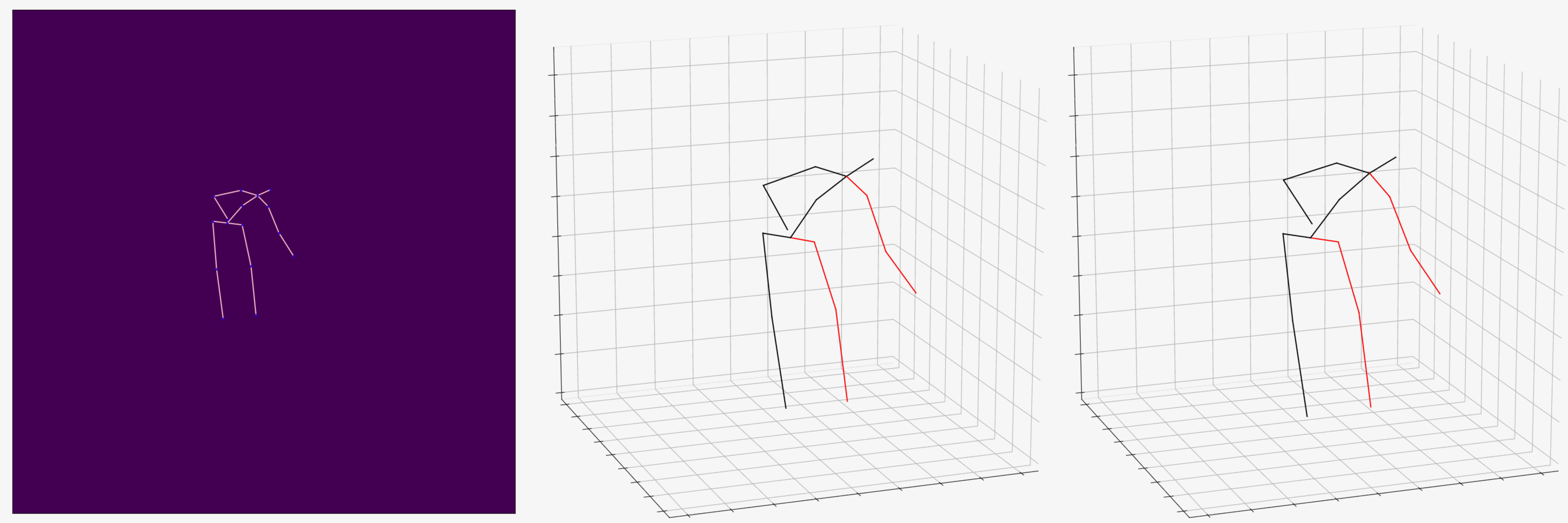} & \includegraphics[width=2.2in]{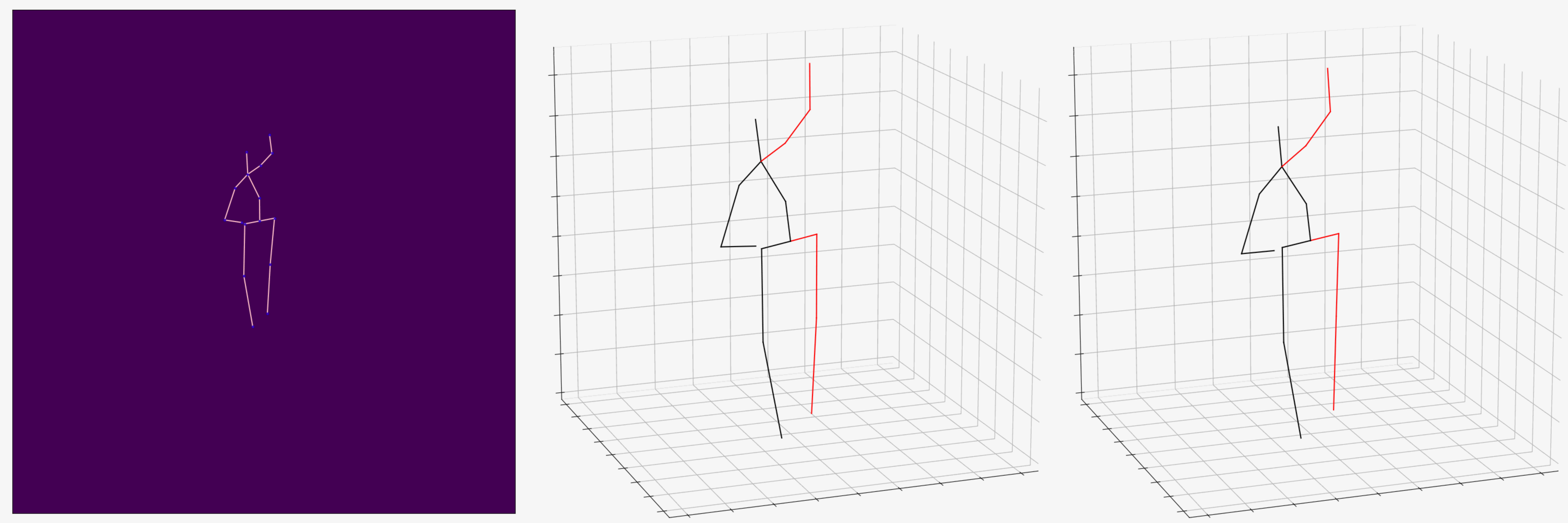}\\
\includegraphics[width=2.2in]{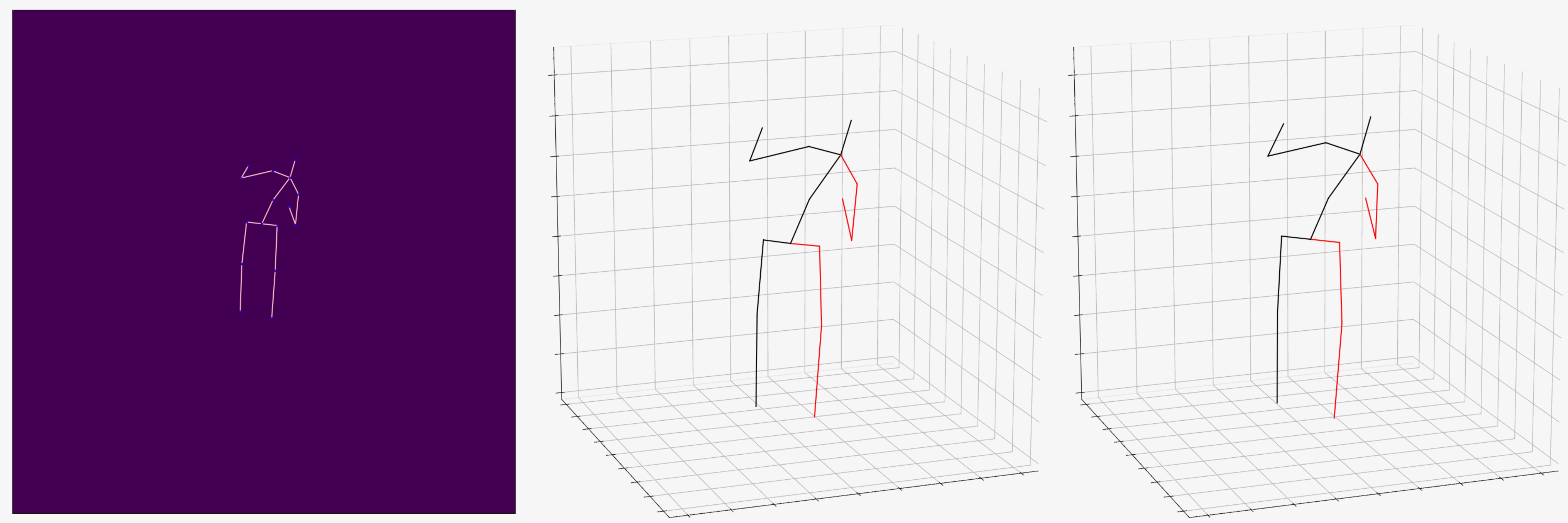} & \includegraphics[width=2.2in]{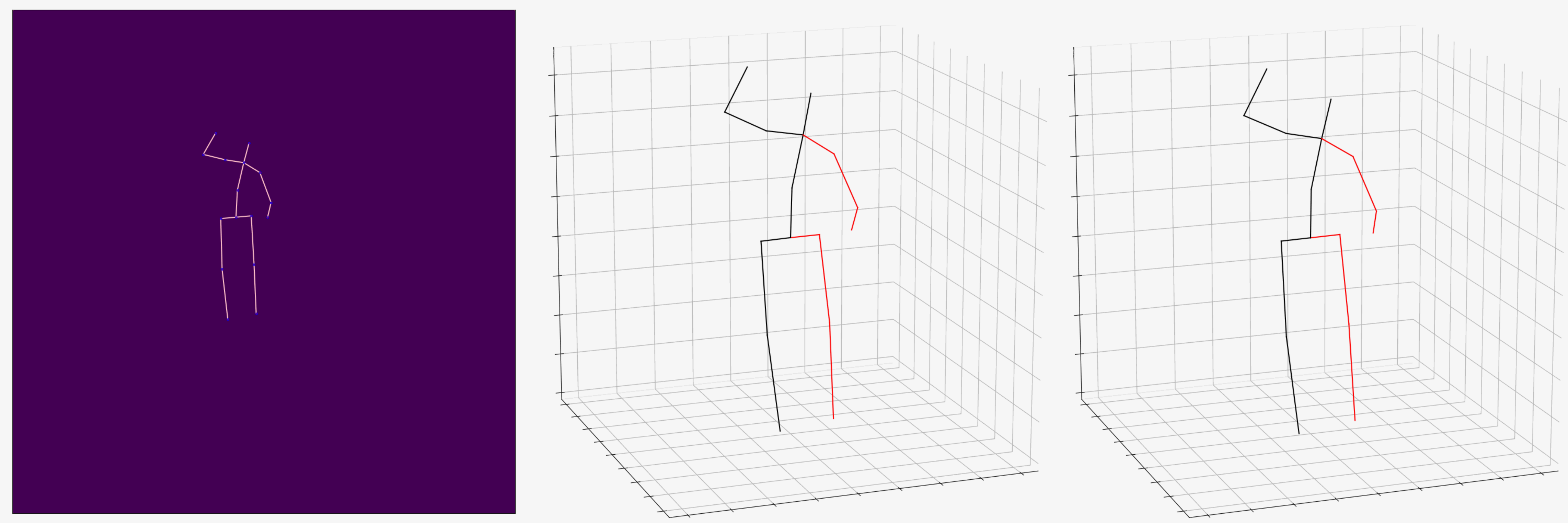}
\end{tabular}
\caption{Qualitative results obtained by our model on the Human3.6M test set.}
\label{Fig:visual}
\end{figure*}

\medskip\noindent\textbf{Qualitative Results.}\quad Figure~\ref{Fig:visual} shows the qualitative results obtained by our model for various actions. Notice that the predictions made by HOIF-Net match perfectly the ground truth, indicating the effectiveness of our proposed approach in tackling the 2D-to-3D pose estimation problem.

\subsection{Ablation study}
\begin{figure}[!htb]
\centering
\setlength{\tabcolsep}{5pt}
\begin{tabular}{cc}
\includegraphics[width=1.8in]{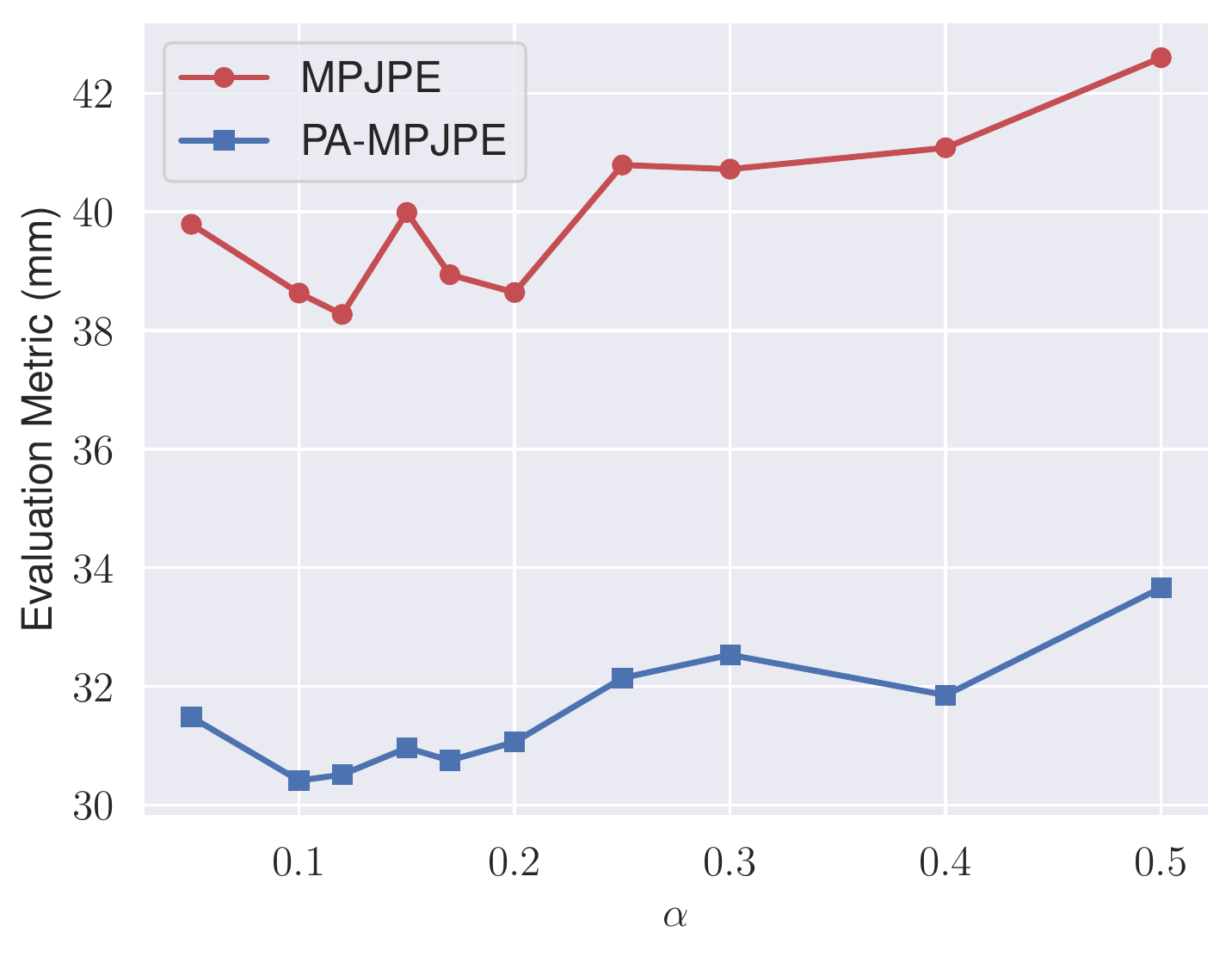} & \includegraphics[width=1.8in]{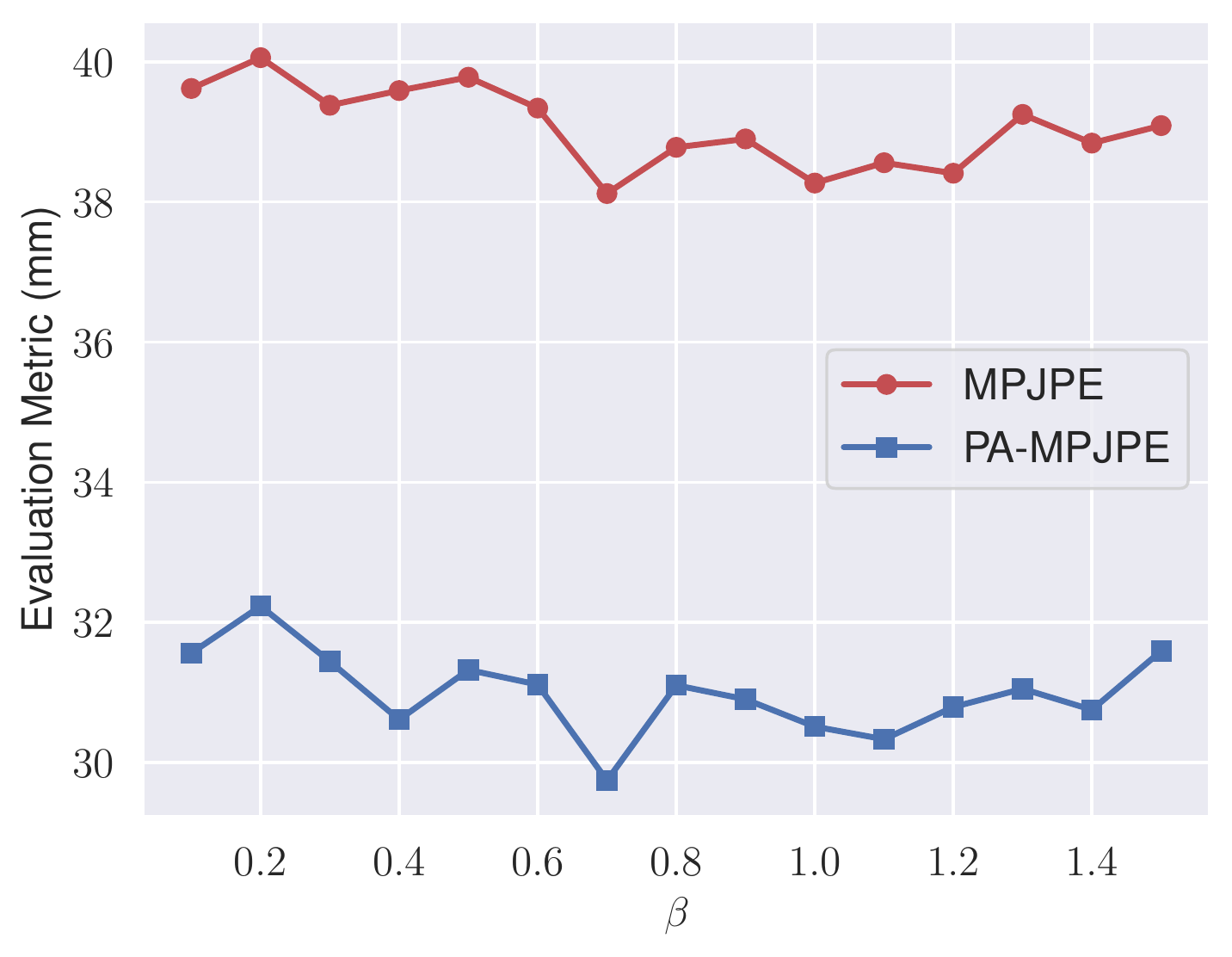}
\end{tabular}
\caption{Parameter sensitivity analysis.}
\label{Fig:alphabeta}
\end{figure}

In our ablation experiments, we use the 2D ground truth as input to our model. We start by investigating the effect of the hyperparameters $\alpha$ and $\beta$ on model performance. We conduct a sensitivity analysis to investigate how the performance of our model changes as we vary these two hyperparameters. In Figure~\ref{Fig:alphabeta} (left), we analyze the effect of $\alpha$ by plotting the error values vs. $\alpha$ for both protocols, where $\alpha$ varies from 0.1 to 0.5, and $\beta$ is set to 1. We can see that our model achieves the lowest error values of MPJPE and PA-MPJPE when $\alpha=0.12$ and $\alpha=0.1$, respectively. In Figure~\ref{Fig:alphabeta} (right), we plot the error values vs. $\beta$ for both protocols by varying the value of $\beta$ from 0.1 to 1.5, and setting the value of $\alpha$ to 0.1. Notice that the best performance is generally achieved when $\beta=0.7$

We also evaluate our method against SemGCN (Zhao \textit{et al.}~\cite{zhao2019semantic}) and high-order GCN (Zou \textit{et al.}~\cite{zou2020high}), which are state-of-the-art GCN-based methods for 2D-to-3D pose estimation, and we report the results in Table~\ref{Tab:baseline}. As can be seen, our approach outperforms both semGCN and High-order GCN under Protocols \#1 and \#2. Under Protocol \#1, our HOIF-Net model outperforms semGCN and high-order GCN by 4.02 mm and 1.4 mm, corresponding to error reductions of 9.54\% and 3.54\%, respectively. Under Protocol \#2, HOIF-Net outperforms semGCN and high-order GCN by 3.79 mm and 1.33 mm, corresponding to error reductions of 11.3\% and 4.28\%, respectively. In addition, our model offers comparable performance as high-order GCN, while using a much smaller number of filters (64 compared to 96) and also the number of learned parameters is reduced by more than half.

\begin{table}[!htb]
\caption{Performance comparison of our model and GCN-based methods.}
\small
\setlength{\tabcolsep}{2.5pt}
\medskip
\centering
\begin{tabular}{l*{7}{c}}
\toprule
Method & Filters & Parameters & MPJPE & PA-MPJPE\\
\midrule
SemGCN~\cite{zhao2019semantic} & 96 & 0.43M & 42.14 & 33.53 \\
%Semgcn w/non-local~\cite{zou2020high} & 96 & 0.43M & 40.78 & 31.46  \\
High-order GCN~\cite{zou2020high}  & 96 & 1.20M & 39.52 &31.07 \\
\midrule
Ours & 96 & 1.20M &\textbf{38.12} &\textbf{29.74} \\
Ours & 64 & \textbf{0.54M} & 39.78 & 31.26\\
\bottomrule

\end{tabular}
\label{Tab:baseline}
\end{table}

\begin{figure}[!htb]
\centering
\setlength{\tabcolsep}{5pt}
\begin{tabular}{cc}
\includegraphics[width=1.9in]{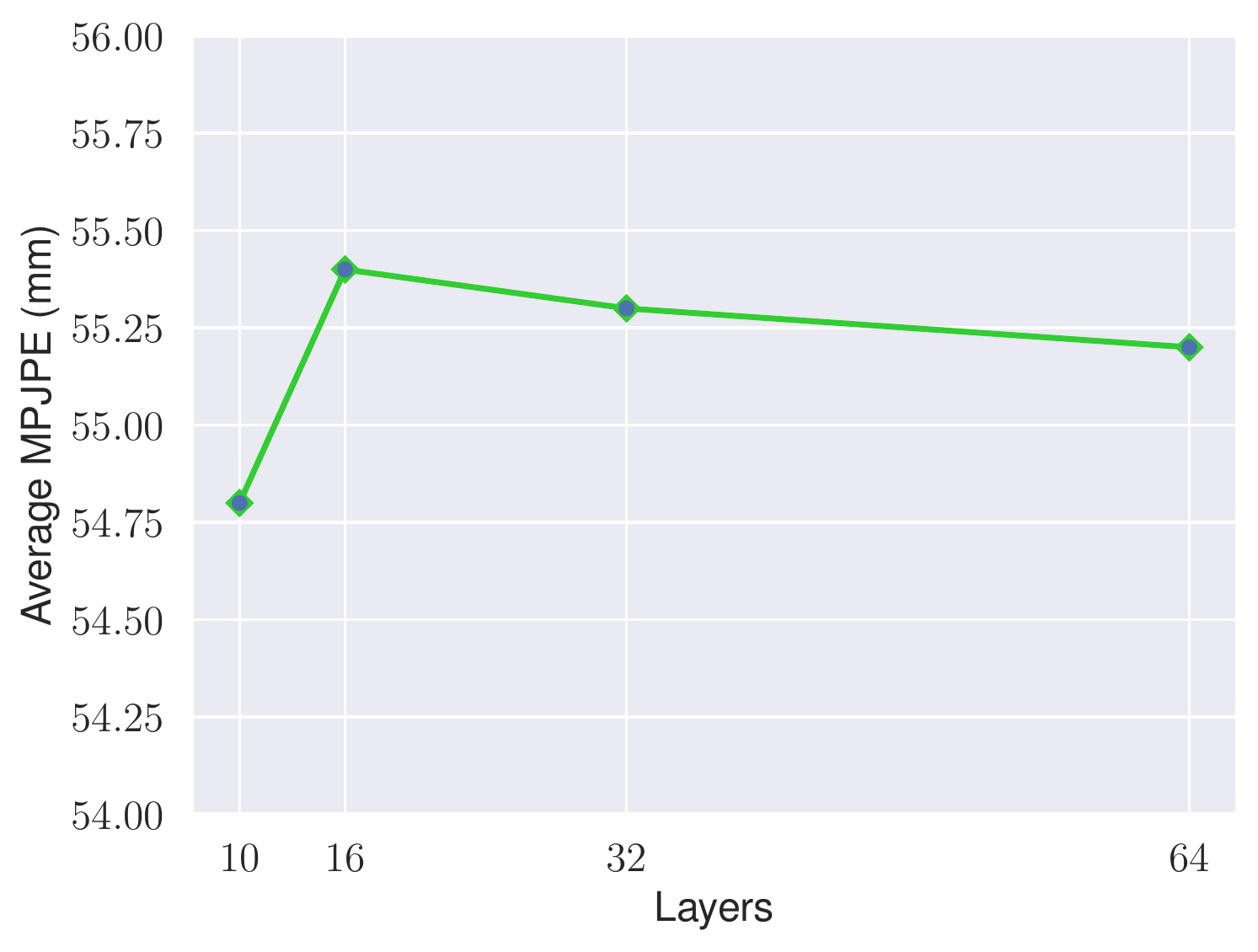} & \includegraphics[width=1.9in]{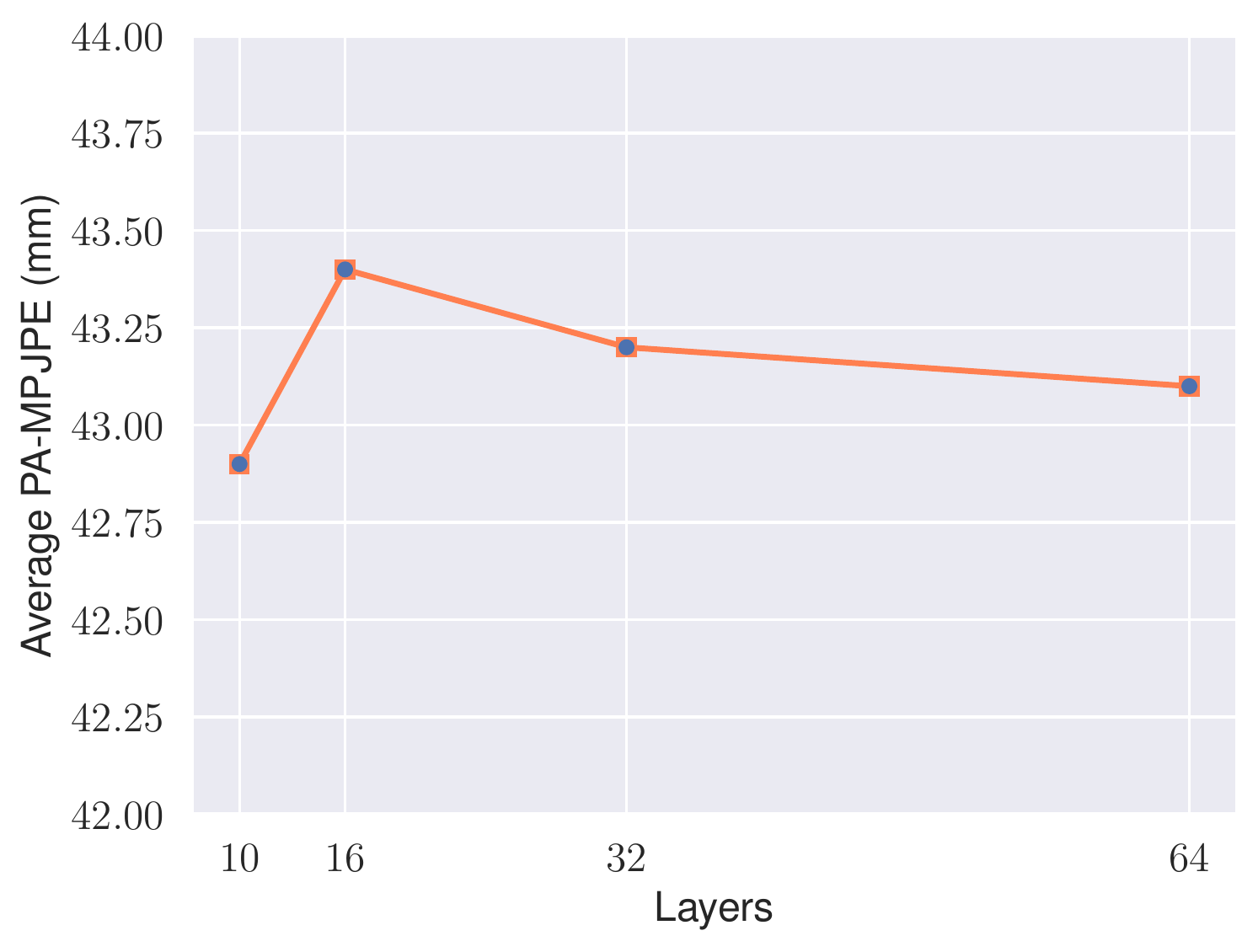}
\end{tabular}
\caption{HOIF-Net's performance with increasing higher-order graph convolutional layers on the Human3.6M dataset.}
\label{Fig:oversmoothing}
\end{figure}

Figure~\ref{Fig:oversmoothing} shows that the performance of the proposed HOIF-Net model on the Human3.6M dataset remains relatively stable as we increase the number of higher-order graph convolutional layers, demonstrating the robustness of our method against oversmoothing.
\section{Conclusion}
In this paper, we proposed a higher-order implicit fairing network with initial residual connections for 3D human pose estimation, with the aim to alleviate the oversmoothing problem in graph convolutional networks, and also to capture long-range dependencies between body joints by enabling the model to aggregate multi-hop neighbors through feature concatenation. Empirical experiments and ablation studies showcase the merits of our model and demonstrate its competitive performance in comparison with state-of-the-art methods for 3D human pose estimation. For future work, we plan to apply the proposed framework to other downstream tasks such as semi-supervised node/graph classification and link prediction.

\bibliographystyle{bmvc2k_natbib}
\bibliography{references}
\end{document}